\documentclass[letterpaper, 10 pt, conference]{class/ieeeconf}  

\IEEEoverridecommandlockouts                              
\usepackage{siunitx}  
\usepackage{graphicx}
 \graphicspath{./figures/robot_demonstration}

\usepackage{amsmath}
\usepackage{amssymb}
\usepackage{latexsym}
\usepackage{url}
\usepackage{cite}
\usepackage{relsize}
\usepackage{multirow}
\usepackage{afterpage}
\usepackage{ifthen}
\usepackage{graphicx}
\usepackage{algpseudocode,algorithm,algorithmicx}
\usepackage{subfigure}
\usepackage{flushend}
\usepackage{epstopdf}
\usepackage{stackengine}
\usepackage{textcomp} 

\usepackage[dvipsnames]{xcolor}

\usepackage{gensymb}
\usepackage{booktabs}
\usepackage{makecell}
\usepackage{hhline}
\usepackage{courier}
\usepackage{lipsum}
\usepackage{tcolorbox}
\newtcbox{\inlinecode}{on line, boxsep=1pt, arc=3pt, colback=black, colframe=black, coltext=white}

\makeatletter
\let\NAT@parse\undefined
\makeatother
\usepackage[bookmarks=false, linkcolor=blue, urlcolor=blue, citecolor=blue]{hyperref} 

\hypersetup{
    colorlinks=true,
    linkcolor=red,
    filecolor=magenta,      
    urlcolor=blue,
    pdfstartview={FitH},
    citecolor =blue
    }

\graphicspath{{./figures/exp}}

\usepackage{xspace}

\title{\LARGE \bf GraspMAS: Zero-Shot Language-driven Grasp Detection with Multi-Agent System}

\author{Quang Nguyen$^{1}$, Tri Le$^{1}$, Huy Nguyen$^{2}$, Thieu Vo$^{3}$, Tung Ta$^{4}$, Baoru Huang$^{5}$, Minh Vu$^{2}$, Anh Nguyen${^5}$
\thanks{$^1$ FPT Software AI Center, Vietnam {\tt quangnv89@fpt.com}}
\thanks{$^2$  Automation \& Control Institute (ACIN), TU Wien, Austria}
\thanks{$^3$ Department of Mathematics, NUS, Singapore}
\thanks{$^4$ Department of Creative Informatics, University of Tokyo, Japan}
\thanks{$^5$ Department of Computer Science, University of Liverpool, UK}
}
\begin{document}

\newtheorem{problem}{Problem}
\newtheorem{lemma}{Lemma}
\newtheorem{theorem}[lemma]{Theorem}
\newtheorem{claim}{Claim}
\newtheorem{corollary}[lemma]{Corollary}
\newtheorem{definition}[lemma]{Definition}
\newtheorem{proposition}[lemma]{Proposition}
\newtheorem{remark}[lemma]{Remark}
\newenvironment{LabeledProof}[1]{\noindent{\it Proof of #1: }}{\qed}

\def\beq#1\eeq{\begin{equation}#1\end{equation}}
\def\bea#1\eea{\begin{align}#1\end{align}}
\def\beg#1\eeg{\begin{gather}#1\end{gather}}
\def\beqs#1\eeqs{\begin{equation*}#1\end{equation*}}
\def\beas#1\eeas{\begin{align*}#1\end{align*}}
\def\begs#1\eegs{\begin{gather*}#1\end{gather*}}

\newcommand{\poly}{\mathrm{poly}}
\newcommand{\eps}{\epsilon}
\newcommand{\e}{\epsilon}
\newcommand{\polylog}{\mathrm{polylog}}
\newcommand{\rob}[1]{\left( #1 \right)} 
\newcommand{\sqb}[1]{\left[ #1 \right]} 
\newcommand{\cub}[1]{\left\{ #1 \right\} } 
\newcommand{\rb}[1]{\left( #1 \right)} 
\newcommand{\abs}[1]{\left| #1 \right|} 
\newcommand{\zo}{\{0, 1\}}
\newcommand{\zonzo}{\zo^n \to \zo}
\newcommand{\zokzo}{\zo^k \to \zo}
\newcommand{\zot}{\{0,1,2\}}
\newcommand{\en}[1]{\marginpar{\textbf{#1}}}
\newcommand{\efn}[1]{\footnote{\textbf{#1}}}
\newcommand{\vecbm}[1]{\boldmath{#1}} 
\newcommand{\uvec}[1]{\hat{\vec{#1}}}
\newcommand{\thv}{\vecbm{\theta}}
\newcommand{\junk}[1]{}
\newcommand{\var}{\mathop{\mathrm{var}}}
\newcommand{\rank}{\mathop{\mathrm{rank}}}
\newcommand{\diag}{\mathop{\mathrm{diag}}}
\newcommand{\tr}{\mathop{\mathrm{tr}}}
\newcommand{\acos}{\mathop{\mathrm{acos}}}
\newcommand{\atantwo}{\mathop{\mathrm{atan2}}}
\newcommand{\SVD}{\mathop{\mathrm{SVD}}}
\newcommand{\quadf}{\mathop{\mathrm{q}}}
\newcommand{\linterp}{\mathop{\mathrm{l}}}
\newcommand{\sgn}{\mathop{\mathrm{sign}}}
\newcommand{\sym}{\mathop{\mathrm{sym}}}
\newcommand{\avg}{\mathop{\mathrm{avg}}}
\newcommand{\mean}{\mathop{\mathrm{mean}}}
\newcommand{\erf}{\mathop{\mathrm{erf}}}
\newcommand{\grad}{\nabla}
\newcommand{\R}{\mathbb{R}}
\newcommand{\defeq}{\triangleq}
\newcommand{\dims}[2]{[#1\!\times\!#2]}
\newcommand{\sdims}[2]{\mathsmaller{#1\!\times\!#2}}
\newcommand{\udims}[3]{#1}
\newcommand{\udimst}[4]{#1}
\newcommand{\com}[1]{\rhd\text{\emph{#1}}}
\newcommand{\ind}{\hspace{1em}}
\newcommand{\argmin}[1]{\underset{#1}{\operatorname{argmin}}}
\newcommand{\floor}[1]{\left\lfloor{#1}\right\rfloor}
\newcommand{\step}[1]{\vspace{0.5em}\noindent{#1}}
\newcommand{\quat}[1]{\ensuremath{\mathring{\mathbf{#1}}}}
\newcommand{\norm}[1]{\left\lVert#1\right\rVert}
\newcommand{\ignore}[1]{}
\newcommand{\specialcell}[2][c]{\begin{tabular}[#1]{@{}c@{}}#2\end{tabular}}
\newcommand*\Let[2]{\State #1 $\gets$ #2}
\newcommand{\algorithmicbreak}{\textbf{break}}
\newcommand{\Break}{\State \algorithmicbreak}
\newcommand{\ra}[1]{\renewcommand{\arraystretch}{#1}}

\renewcommand{\vec}[1]{\mathbf{#1}} 

\algdef{S}[FOR]{ForEach}[1]{\algorithmicforeach\ #1\ \algorithmicdo}
\algnewcommand\algorithmicforeach{\textbf{for each}}
\algrenewcommand\algorithmicrequire{\textbf{Require:}}
\algrenewcommand\algorithmicensure{\textbf{Ensure:}}
\algnewcommand\algorithmicinput{\textbf{Input:}}
\algnewcommand\INPUT{\item[\algorithmicinput]}
\algnewcommand\algorithmicoutput{\textbf{Output:}}
\algnewcommand\OUTPUT{\item[\algorithmicoutput]}

\maketitle
\thispagestyle{empty}
\pagestyle{empty}

\begin{abstract}
Language-driven grasp detection has the potential to revolutionize human-robot interaction by allowing robots to understand and execute grasping tasks based on natural language commands. However, existing approaches face two key challenges. First, they often struggle to interpret complex text instructions or operate ineffectively in densely cluttered environments. Second, most methods require a training or fine-tuning step to adapt to new domains, limiting their generation in real-world applications. In this paper, we introduce GraspMAS, a new multi-agent system framework for language-driven grasp detection. GraspMAS is designed to reason through ambiguities and improve decision-making in real-world scenarios. Our framework consists of three specialized agents: \texttt{Planner}, responsible for strategizing complex queries; \texttt{Coder}, which generates and executes source code; and \texttt{Observer}, which evaluates the outcomes and provides feedback. Intensive experiments on two large-scale datasets demonstrate that our GraspMAS significantly outperforms existing baselines. Additionally, robot experiments conducted in both simulation and real-world settings further validate the effectiveness of our approach. Our project page is available at \href{https://zquang2202.github.io/GraspMAS}{https://zquang2202.github.io/GraspMAS}.
\end{abstract}


\vspace{-3pt}
\section{INTRODUCTION} \label{Sec:Intro}
Grasp detection is a foundational aspect of robotic manipulation, enabling robots to determine feasible grasping configurations for effectively interacting with objects across various environments~\cite{kleeberger2020survey, caldera2018review}. Traditional grasp detection approaches~\cite{redmon2015real} have primarily emphasized securing the stability of identified grasp poses, frequently overlooking the integration of human intent in the process~\cite{nguyen2016preparatory, depierre2018jacquard,fang2020graspnet, wang2022transformer}. Recently, several works have integrated natural language into robotic grasping to enhance intuitive and flexible human-robot interaction, allowing robots to comprehend and execute context-aware instructions~\cite{vuong2024language,tziafas2023language}. However, most existing grasp detection methods rely on a resource-intensive training process, which limits their adaptability to real-world scenarios, especially when faced with unseen data.

Recently, progress in large-scale foundation models~\cite{brown2020language, gpt, touvron2023llama, liu2024deepseek} has fueled significant growth in robotics research, particularly in enabling intuitive natural language interaction. Researchers have developed zero-shot approaches for different robotic tasks such as planning~\cite{ahn2022can, singh2023progprompt}, manipulation~\cite{liang2023code, zeng2022socratic}, and navigation~\cite{yu2023l3mvn, lin2022adapt}. However, several existing works focus on long manipulation tasks while overlooking the foundational aspect of grasping. Some studies~\cite{vo2024language, shridhar2022cliport, xu2023joint} design neural architectures to fuse visual and textual inputs, while others~\cite{vuong2024language,nguyen2024language, nguyen2024lightweight} adopt generative frameworks trained on extensive multimodal datasets~\cite{vuong2023grasp}. Recent works utilize foundation models to enable zero-shot adaptation in robotics~\cite{tang2023graspgpt, tziafas2024towards,qian2024thinkgrasp}. However, these methods still exhibit certain limitations. First, prior approaches lack the compositional reasoning needed to interpret open-ended language commands. For instance, a query like ``Grasp the second bottle from the left" requires the model to spatially localize objects, count ordinal positions, and resolve directional references, while a command like ``I need something to cut" demands inferring functional object properties (e.g., identifying knives or scissors as tools for cutting). Second, their reliance on end-to-end monolithic neural architectures necessitates training on large-scale datasets, a process that is resource-intensive and dataset-dependent. Furthermore, these models often lack generalization capabilities, requiring additional domain-specific training to achieve reliable performance when deployed across diverse scenarios~\cite{fang2020graspnet}.

\begin{figure}[t]
    \centering
    \subfigure[Traditional methods]{
        \includegraphics[width=0.4\columnwidth]{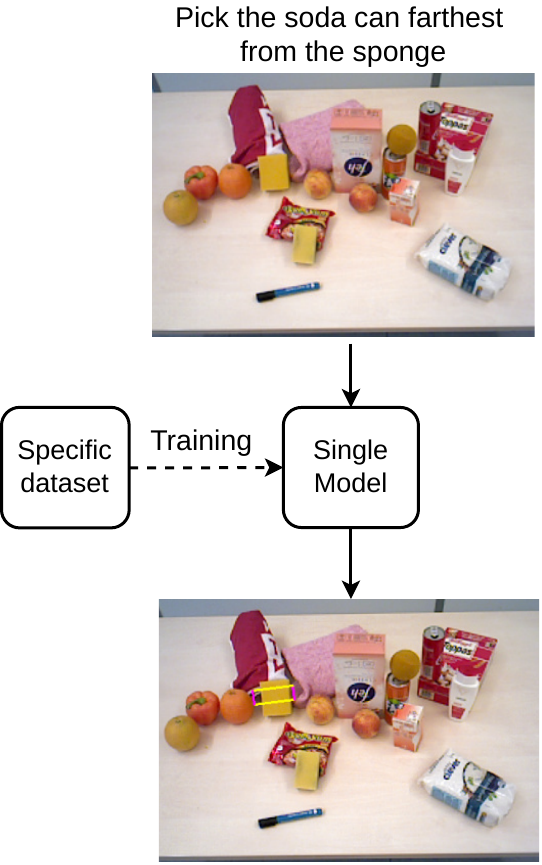}
    } 
    \hspace{2pt}
    \subfigure[Our method]{
        \includegraphics[width=0.4\columnwidth]{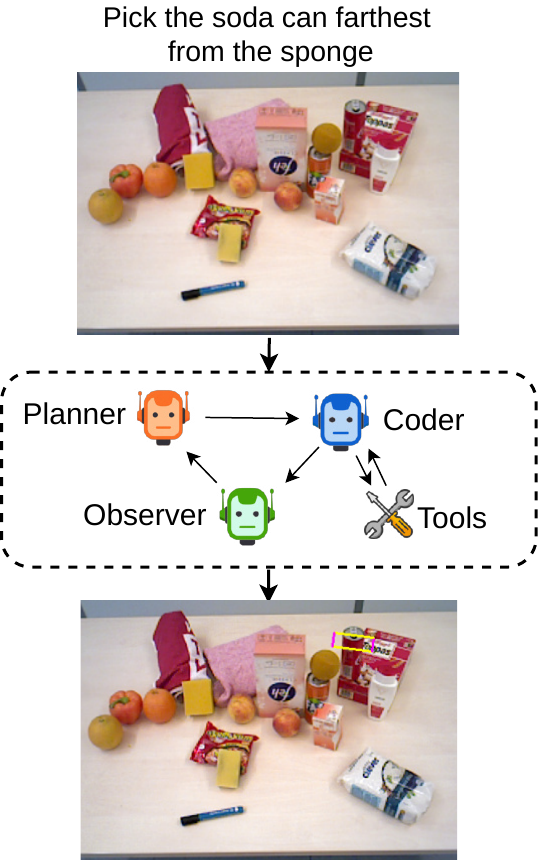}
    }
    \vspace{5pt}
    \caption{Comparison between the traditional methods (a) and our approach (b) for language-driven grasp detection. The traditional method relies on a single trainable network, which may struggle with complex queries. In contrast, our approach leverages multi-agent collaboration, enabling more flexible and accurate grasp detection.}
    \label{fig: intro}
\end{figure}
\vspace{-2ex}
To tackle real-world diverse scenarios, multi-agent systems that use compositional approaches~\cite{gupta2023visual,lu2024chameleon,suris2023vipergpt} have emerged as powerful strategies to reason and make decisions collaboratively using several specialized subsystems (i.e., agents). Multi-agent systems have shown success in domains such as visual question answering~\cite{gupta2023visual, suris2023vipergpt}, video analysis~\cite{fan2024videoagent}, and robotics~\cite{li2024robocoder, qin2024mp5, wang2023describe}, where they integrate large language models and visual foundation models with minimal fine-tuning. In such frameworks, large language models usually serve as planners, code generators, or reasoning engines, while visual foundation models enable robust visual perception, fostering structured reasoning and task planning to enhance cross-domain generalization. Inspired by this paradigm, we propose a modular framework for language-conditioned grasp detection, designed to connect semantic language understanding with precise robotic physical execution in diverse real-world scenarios.

In this paper, we propose GraspMAS, a zero-shot grasp detection method based on multi-agent systems. Our approach is a training-free, plug-and-play framework that leverages multi-agent collaboration and re-uses existing foundation vision language models (VLMs) to achieve robust language-driven grasp detection. Our framework comprises three core agents: \texttt{Planner}, \texttt{Coder}, and \texttt{Observer}. The \texttt{Planner} serves as a reasoning agent that generates detailed instructional plans. The \texttt{Coder} then translates these plans into executable Python code, dynamically interfacing with VLMs to detect objects and resolve spatial relation ambiguities. Finally, the \texttt{Observer} processes intermediate results from the executed code and generates a summary report, which is fed back to the \texttt{Planner}. This feedback mechanism enables the \texttt{Planner} to refine and adjust its instructional plans accordingly. In summary, our key contributions are:
\begin{itemize}
    \item We present GraspMAS, a new multi-agent approach for zero-shot language-driven grasp detection.
    \item We validate our method across diverse settings, including benchmark datasets, simulated environments, and real robotic experiments. The results show that our method significantly outperforms other recent baselines.
\end{itemize}

\section{Related Work} \label{Sec:rw}
\textbf{Language-driven Grasp Detection.} Traditional robotic grasping systems rely on analytical approaches that model object geometry and contact forces~\cite{maitin2010cloth, domae2014fast, roa2015grasp}, or on convolutional neural networks (CNNs) trained on labeled grasp datasets~\cite{lenz2015deep, pinto2016supersizing, kumra2020antipodal, depierre2018jacquard, vuong2023grasp}. While effective in controlled settings, these methods lack contextual awareness and cannot interpret natural language instructions, limiting their adaptability in human-centric environments. Recent advancements address this gap by integrating language understanding with vision systems, enabling robots to localize and grasp objects specified through natural language~\cite{vuong2024language, nguyen2024lightweight, lu2023vl, xu2023joint, cheang2022learning, nguyen2024language}. A key innovation involves aligning textual and visual embeddings in a shared latent space, allowing models to correlate linguistic cues with object regions for grasp prediction. For example, Tziafas \textit{et al.}~\cite{tziafas2023language} leverage linguistic references to synthesize grasps in cluttered scenes, while Chen \textit{et al.}~\cite{chen2021joint} fuse multimodal features to predict 2D grasp boxes from RGB images. Large language models (LLMs) further enhance this paradigm by enabling reasoning over indirect or abstract query. Jin \textit{et al.}~\cite{jin2024reasoning}, for instance, employ LLMs to infer grasp poses from verbal commands that lack explicit object references. Despite these advances, previous methods rely on simple text prompts and struggle with complex language, and often requiring additional fine-tuning. In contrast, our approach employs a multi-agent framework that re-uses existing foundation models to effectively interpret diverse language instructions without extra training.

\textbf{Multi-Agent Systems for Robotics.} LLM-based agents have recently gained significant attention from researchers and are considered a key step toward artificial general intelligence~\cite{bubeck2023sparks}. With advancements in task reasoning, decomposition, and instruction following, these agents have demonstrated remarkable success across various domains, including visual question answering~\cite{gupta2023visual, suris2023vipergpt, ke2024hydra, hu2024visual, yuan2023craft}, video analysis~\cite{fan2024videoagent}, biological research~\cite{jin2024genegpt}, and embodied AI robotics~\cite{li2024robocoder, zhang2023building, qin2024mp5, wang2023describe}. HuggingGPT~\cite{shen2024hugginggpt} leverages LLMs as planners to coordinate expert AI models for solving complex AI tasks. VisProg~\cite{gupta2023visual} and ViperGPT~\cite{suris2023vipergpt} utilize Codex to generate Python code that interacts with APIs for visual reasoning tasks without requiring additional training. In~\cite{zhang2023building}, the authors present CoELA, a cooperative embodied agent that integrates perception, memory, and execution modules, enabling efficient collaboration with other agents and humans. In this paper, we propose a new multi-agent framework for language-driven grasp detection. Our framework integrates reasoning capabilities to enhance zero-shot detection performance without requiring additional training.

\textbf{Zero-shot learning.} Zero-shot methods are vital for real-world applications, enabling models to recognize unseen categories without direct supervision. Recent approaches~\cite{liu2021learning, javed2024cplip, xie2022zero} align visual and pretrained text embeddings to facilitate zero-shot learning but often face challenges in spatial reasoning. Recent methods~\cite{wang2024qwen2, tziafas2024towards,singh2024malmm} leverage VLMs to enhance reasoning and semantic understanding, boosting zero-shot performance. SoM~\cite{yang2023set} employs visual prompting with symbolic markers for diverse vision tasks, while ShapeGrasp~\cite{li2024shapegrasp} introduces a geometric composition framework for zero-shot task-oriented grasp detection. Agentic methods~\cite{suris2023vipergpt, ke2024hydra, fan2024videoagent} further expand zero-shot learning paradigms. Building on these, our approach proposes a multi-agent framework to enhance reasoning and adaptability.

\section{Multi-agent Grasp Detection} \label{Sec:method}
\begin{figure*}[ht]
    \centering
    \includegraphics[width=\textwidth]{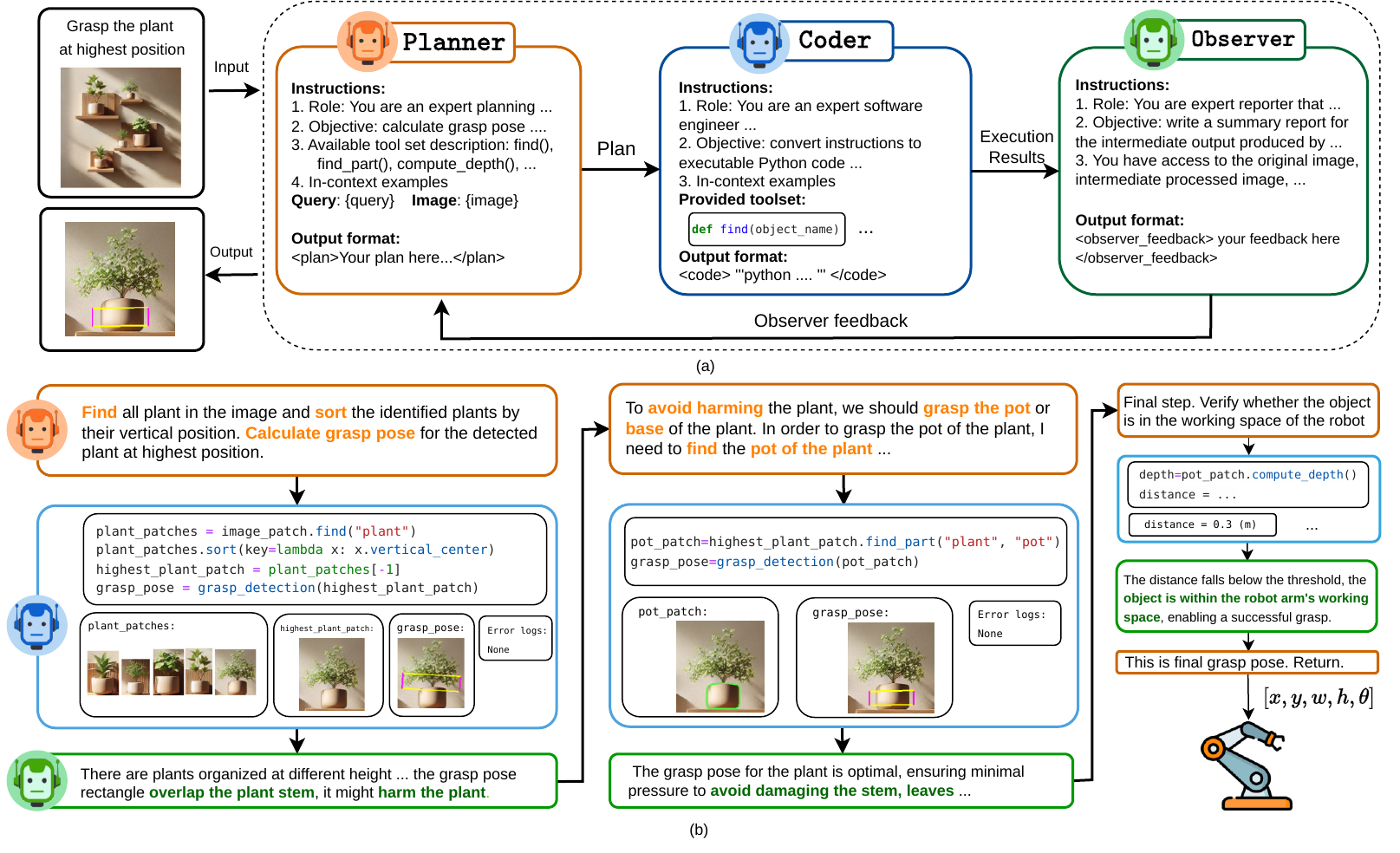}
    \vspace{-2pt}
    \caption{(a) An overview of our GraspMAS framework, a multi-agent system for zero-shot language-driven grasp detection. GraspMAS consists of three agents: \texttt{Planner} (\textcolor{BurntOrange}{orange box}) – controls the reasoning flow; \texttt{Coder} (\textcolor{Blue}{blue box}) – executes instructions using the provided toolset; and \texttt{Observer} (\textcolor{Green}{green box}) – evaluates results and provides feedback to the \texttt{Planner}; (b) A successful example demonstrating how our framework enables reasoning grasp detection by leveraging multi-agents knowledge and commonsense understanding.}
    \label{fig: method}
\end{figure*}
\subsection{Overview}
Given an input RGB image and a free-form text query, our approach aims to detect the feasible grasping pose that aligns with the textual input. Similar to~\cite{depierre2018jacquard}, the grasp pose is represented as a rectangle defined by five parameters: the center coordinate $(x, y)$, the width and height $(w, h)$ of the rectangle, and the rotational angle $\theta$, which specifies its orientation relative to the horizontal axis of the image. The pipeline of our proposed method is illustrated in Fig.~\ref{fig: method}. The framework comprises three agents: \textit{i)} The \texttt{Planner}, which analyzes input images, task requirements, and extracted visual cues to generate sequential grasping instructions; \textit{ii)} The \texttt{Coder} translates these instructions into executable Python code; and \textit{iii)} The \texttt{Observer} monitors intermediate results during the reasoning process to evaluate the reliability of extracted visual cues and intermediate outputs. Through a closed-loop interaction among these three agents, the system dynamically refines its reasoning, enabling the efficient translation of high-level commands into accurate grasp poses. This iterative mechanism ensures robust adaptation to complex task constraints and environmental uncertainties.

\subsection{Planner Agent} 
Inspired by recent advancements in LLM-based agent frameworks~\cite{ke2024hydra, yang2023mm}, we construct a \texttt{Planner} agent as the brain of the system. Its primary goal is to determine the detailed grasping strategy. Additionally, the \texttt{Planner} agent actively communicates with other agents in the system to exchange critical information. It receives feedback from the \texttt{Observer} agent to iteratively refine and improve its plan, ensuring deep understanding of the environment and enhancing the overall grasping performance. At each time, this planning agent takes three types of inputs:
\begin{itemize}
    \item \textbf{Instructions}: We provide a specified role, clear objective, description of the available toolsets, and in-context examples. The objective of the \texttt{Planner} is to locate the query object, compute the grasp pose, and verify whether the grasp pose is feasible for execution. Additionally, an abstract description of the provided tool usage is included, ensuring that the agent can generate a comprehensive plan within the constraints of the available toolset (defined in Table~\ref{tab: func}). 
    \item \textbf{User inputs}: We take both an image and a text query as inputs to the \texttt{Planner} agent instead of only using text prompt as input like in~\cite{gupta2023visual, ke2024hydra}. This approach enhances specificity for individual data points and improves analysis by identifying key elements and critical visual features for a concrete plan.
    \item \textbf{Feedback from \texttt{Observer} agent}: To assist the \texttt{Planner} in making informed decisions and refining the reasoning process, the \texttt{Observer} agent provides textual feedback of intermediate results (e.g., object crops and grasp poses). The \texttt{Planner} combines all the feedback with the original instruction to create a new plan for the next step.
\end{itemize}

We implement the \texttt{Planner} using GPT-4~\cite{gpt}, leveraging its advanced reasoning and planning capabilities to enhance grasp detection. The \texttt{Planner} generates a step-by-step plan specifying the next actions, which the \texttt{Coder} then executes. We implement the \texttt{Planner} to handle two subtasks:  

\begin{itemize}
    \item \textbf{Determining the grasp pose for the target object based on the user’s query.} This process may involve multiple iterations to resolve spatial relationships between objects and clarify text ambiguities. The grasp pose must be physically feasible, ensuring safe and effective interaction. For instance, the robot should grasp the pot of a plant rather than the plant itself to avoid damage (Fig.~\ref{fig: method} (b)).  

    \item \textbf{Verifying whether the object is within the robot's workspace.} Unlike prior methods~\cite{vuong2023grasp, vuong2024language} that compute grasp poses without considering physical execution constraints, the \texttt{Planner} in our approach ensures practical feasibility with real robots by confirming that the object is physically reachable before proceeding.  
\end{itemize}

\begin{figure}[t]
    \centering
    \includegraphics[width=\linewidth]{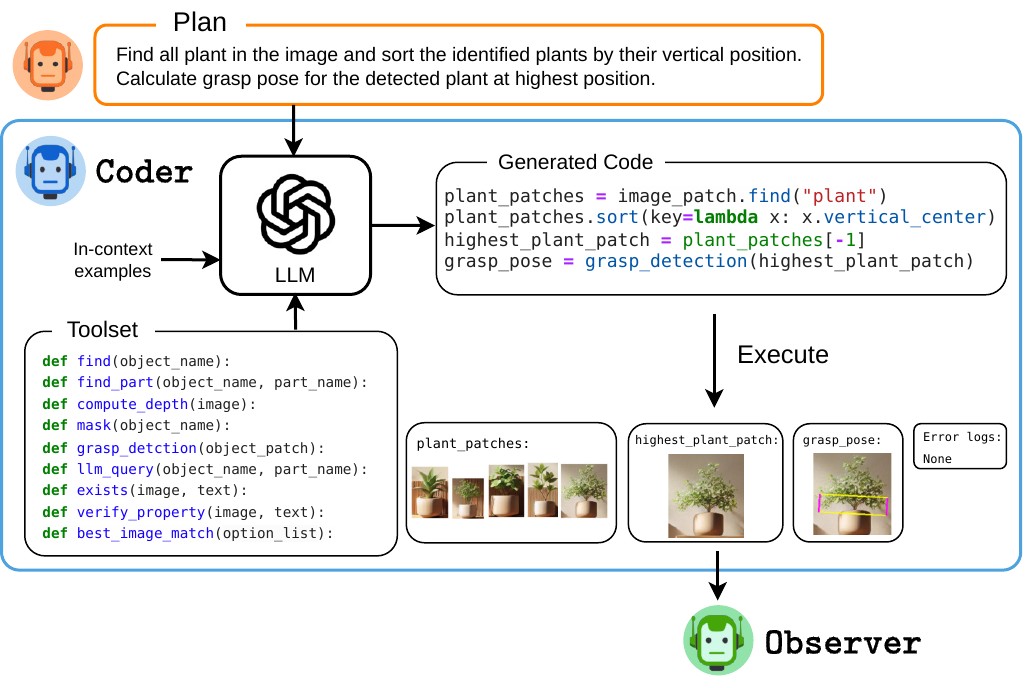}
    \caption{Diagrammatic illustration of the \texttt{Coder} agent.}
    \label{fig: coder}
\end{figure}

\subsection{Coder Agent} 
The detailed design of the \texttt{Coder} agent is illustrated in Fig.~\ref{fig: coder}. The \texttt{Coder} serves as the operator of the system, translating high-level reasoning into concrete actions. The \texttt{Coder} receives plan instructions from the \texttt{Planner}, along with a structured prompt designed to generate Python code. The prompt includes guiding text, predefined Python functional tools, and in-context examples of how to use these functions correctly. The generated Python code is compiled and executed by a Python interpreter. All variable values, such as processed images and logs, are sent to the \texttt{Observer}. A key aspect of our design is that \textit{if an error occurs during code execution, the process is not interrupted}. Instead, error logs are returned to the \texttt{Observer}, which summarizes them and sends the information back to the \texttt{Planner} for further refinement.

\begin{table}[ht]
\centering
\caption{Toolset summarization}
\vspace{10pt}
\resizebox{\linewidth}{!}{
\begin{tabular}{lll}
\toprule
\textbf{Tool} & \textbf{Uses} &\textbf{Model} \\ \midrule
\texttt{find} & Detect object given object name& GrDINO~\cite{liu2024grounding} \\
\texttt{find\_part} & Detect a part of an object given part name& VLPart~\cite{sun2023going} \\
\texttt{grasp\_detection} & Calculate grasp pose given target object & RAGT-3/3~\cite{cao2023nbmod} \\
\texttt{exists} & Check if object exists in the image &  BLIP2~\cite{li2023blip} \\
\texttt{verify\_property} & Verify property of an object (color, shape)& BLIP2~\cite{li2023blip} \\
\texttt{best\_image\_match} & Returns the patch most likely to contain the content & BLIP2~\cite{li2023blip} \\
\texttt{compute\_depth} & Return a median depth of an image& MiDaS~\cite{ranftl2020towards} \\
\texttt{masks} & Return mask of an object given object name & SAM~\cite{kirillov2023segment} \\ 
\texttt{llm\_query} & Ask external information about the image & GPT-4~\cite{gpt} \\
\bottomrule
\end{tabular}}
\label{tab: func}
\end{table}

Unlike other approaches that include a vast array of tools that often result in ambiguity in selection and a complicated usage pipeline~\cite{yuan2023craft}, our system offers a minimal but sufficient set of tools (i.e., core Python functions) specifically designed for language-driven grasp detection. These tools cover key tasks such as referring object detection, depth estimation, and grasp detection. Overall, our system features 9 tool functions, which we detail along with their applications and corresponding foundation models in Table~\ref{tab: func}.

\subsection{Observer Agent}
To enhance the system's robustness and flexibility in diverse situations, we employ an \texttt{Observer} that manages the execution results and provides feedback to the \texttt{Planner}. The feedback from \texttt{Observer} is crucial for ensuring coordinated adaptation, allowing the \texttt{Planner} to adjust its plans based on the success or challenges encountered during grasp detection. The \texttt{Observer} is a multi-modal agent that takes two inputs: \textit{(i)} the visualized images generated by the \texttt{Coder} (e.g., the cropped image of a plant, grasp pose rectangle as in Fig.~\ref{fig: method} (b), and \textit{(ii)} the code execution error logs (if any) of the \texttt{Coder} agent. Given these two inputs, the \texttt{Observer} does the reasoning and outputs textual feedback and sends it to the \texttt{Planner}. Specifically, we explicitly prompt the \texttt{Observer} to evaluate the generated grasp pose to assess potential risks, such as harm to objects or humans or the likelihood of collisions. After receiving the feedback, the \texttt{Planner} then refines the plan and sends it to the \texttt{Coder} for generating new execution code. The loop between three agents allows us to generate valid and feasible grasp poses after one or several reasoning loops. In practice, we use GPT-4o~\cite{gpt} as the \texttt{Observer} as it possesses universal knowledge and multi-modal processing capabilities.


\section{Experiments} \label{Sec:exp}
We evaluate the effectiveness of our method using two language-driven grasping datasets, Grasp-Anything++~\cite{vuong2024language} and OCID-VLG~\cite{tziafas2023language}. We then conduct grasping experiments with both a simulated and a physical robot. Moreover, we showcase our method’s ability in complex in-the-wild experiments. Lastly, we discuss the limitation of our method and highlight open questions for future research.
\subsection{Zero-shot Language-driven Grasp Detection}
\textbf{Dataset.} To evaluate our method, we use two language-driven grasp detection datasets: GraspAnything++~\cite{vuong2024language} and OCID-VLG~\cite{tziafas2023language}. GraspAnything++ comprises 1 million images with textual descriptions and over 3 million objects, synthesized using foundational models. OCID-VLG focuses on cluttered tabletop environments, containing 1,763 scenes and 31 graspable objects. These datasets serve as a robust benchmark, allowing us to evaluate our approach across a variety of complex scenarios.

\begin{figure*}[ht]
    \centering
    \includegraphics[width=\linewidth]{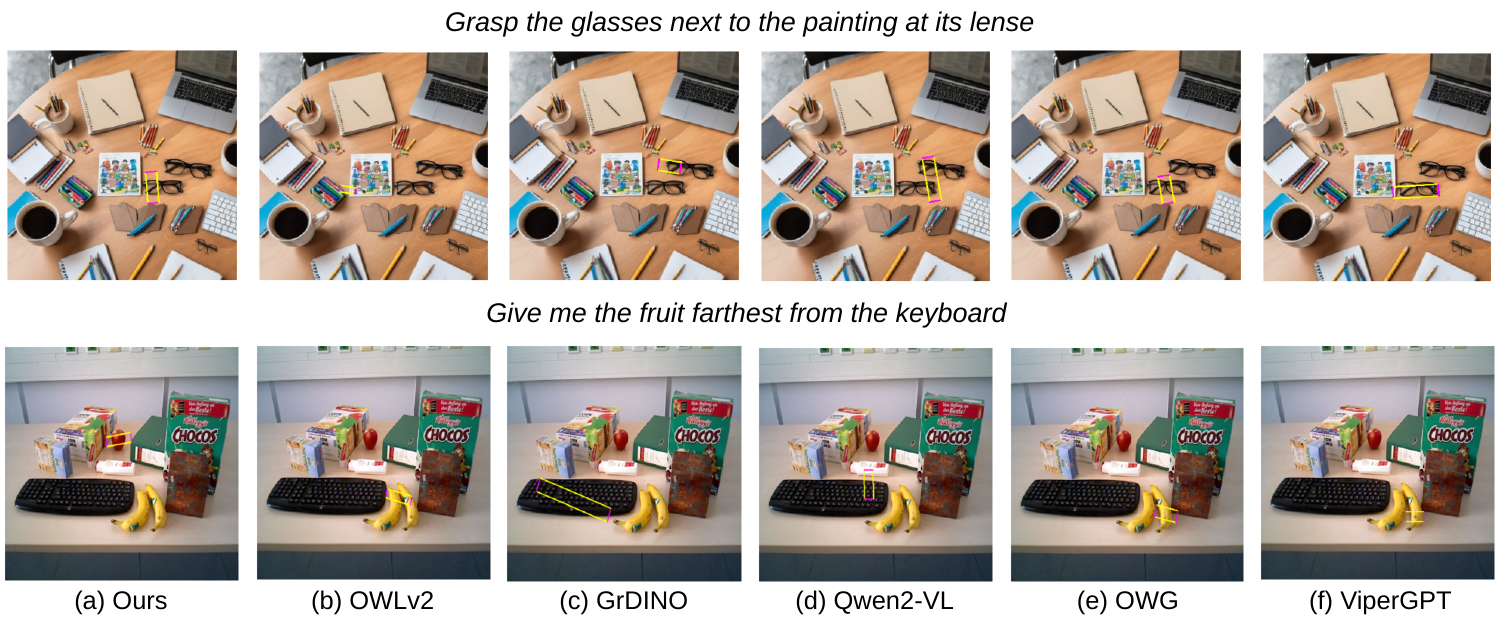}
    \vspace{-1ex}
    \caption{Qualitative comparison with other methods.}
    \label{fig: compare_methods}
\end{figure*}

\textbf{Evaluation metrics.} We evaluate our method using the success rate metric defined in~\cite{depierre2018jacquard}, which considers a grasp successful if its IoU with the ground truth exceeds 0.25 and its orientation differs by no more than 30 degrees. For the GraspAnything++ dataset, we follow the test split defined in~\cite{vuong2024language} and report the success rate using Harmonic Mean metric. Meanwhile, for the OCID-VLG dataset, we report the success rate on its test split, which consists of 17.7K samples. We also report the inference time (in seconds) for all baselines, evaluated on an NVIDIA RTX 4080 GPU.

\textbf{Baselines.} In the zero-shot setting, we compare our approach against large-scale detection models, including GroundingDINO~\cite{liu2024grounding}, OWLv2~\cite{minderer2024scaling}, and QWEN2-VL~\cite{wang2024qwen2}. For these models, we integrate RAGT~\cite{cao2023nbmod}, a grasp detection model, to generate grasp poses based on object grounding. Additionally, we compare our method with OWG~\cite{tziafas2024towards}, an LLM-based approach for open-world grasping, and ViperGPT~\cite{suris2023vipergpt}, a tool-augmented framework for visual inference. To ensure a fair comparison, we equip ViperGPT with the same toolset as our approach, as outlined in Table~\ref{tab: func}. 

\begin{table}[h]
\caption{Zero-shot language-driven grasp detection results}
\centering
\vspace{10pt}
\setlength{\tabcolsep}{0.25 em}
\resizebox{\linewidth}{!}{
\begin{tabular}{lcccc}
\toprule
\textbf{Method} & \textbf{Type} & \textbf{OCID-VLG}~\cite{tziafas2023language} & \textbf{GraspAnything++}~\cite{vuong2024language} & \textbf{Inference (s)}\\
\midrule
OWLv2~\cite{minderer2024scaling} + RAGT~\cite{cao2023nbmod} & \multirow{3}{*}{E2E} & 0.22 \textcolor{OliveGreen}{(+0.40)} & 0.24 \textcolor{OliveGreen}{(+0.44)} & $0.37^{\pm 0.002}$\\
GrDINO~\cite{liu2024grounding} + RAGT~\cite{cao2023nbmod} & &0.27 \textcolor{OliveGreen}{(+0.35)} & 0.33 \textcolor{OliveGreen}{(+0.35)} & $\mathbf{0.32^{\pm 0.002}}$\\
QWEN2~\cite{wang2024qwen2} + RAGT~\cite{cao2023nbmod} & & 0.41 \textcolor{OliveGreen}{(+0.21)} & 0.48 \textcolor{OliveGreen}{(+0.20)} & $1.13^{\pm 0.200}$\\ \midrule
OWG~\cite{tziafas2024towards} & \multirow{3}{*}{Com} & 0.53 \textcolor{OliveGreen}{(+0.09)} & 0.42 \textcolor{OliveGreen}{(+0.26)} & $3.35^{\pm 0.820}$\\
ViperGPT~\cite{suris2023vipergpt}& & 0.44 \textcolor{OliveGreen}{(+0.18)} & 0.57 \textcolor{OliveGreen}{(+0.11)} & $1.04^{\pm 0.300}$ \\
GraspMAS (ours) & & \textbf{0.62} & \textbf{0.68} & $2.12^{\pm 0.500}$\\ 
\bottomrule
\end{tabular}}
\label{tab: zero_shot_compare}
\end{table}

\textbf{Results.} Table~\ref{tab: zero_shot_compare} presents a comparison of our method with baseline approaches in the zero-shot setting. The results demonstrate that our method, leveraging a closed-loop reasoning mechanism, significantly outperforms the others by a large margin. Additionally, we observe that all compositional approaches (Com) achieve higher performance than end-to-end (E2E) baselines. We note that the inference time of our method and other compositional approaches (OWG~\cite{tziafas2024towards}, ViberGPT~\cite{suris2023vipergpt}) is longer than the E2E methods due to the time required for reasoning between multiple agents.

\subsection{Comparison with Supervised Methods}
In this experiment, we compare our zero-shot method with supervised methods trained on GraspAnything++~\cite{vuong2024language} including CLIP-Fusion~\cite{xu2023joint}, CLIPort~\cite{shridhar2022cliport}, LGD~\cite{vuong2024language}, MaskGrasp~\cite{van2024language}, LLGD\cite{nguyen2024lightweight}, and GraspMamba~\cite{nguyen2024graspmamba}. Our evaluation metric is the success rate as defined in~\cite{vuong2023grasp}. Table~\ref{tab: compare_sup} shows that, despite not being trained on grasp-specific datasets, our method surpasses these supervised approaches, demonstrating its strong generalization ability.

\subsection{Qualitative Results}
\textbf{Qualitative comparison.} Fig.~\ref{fig: compare_methods} presents a qualitative comparison of our method against baseline approaches. The results demonstrate that even in cluttered scenes and with complex text prompts, our method successfully generates plausible grasp poses. In contrast, large detection models (OWLv2, GroundingDINO, and QWEN2-VL) struggle with complex prompts, while OWG and ViperGPT lack the reasoning and semantic understanding of the object.

\begin{figure*}[ht]
    \centering
    \includegraphics[width=\linewidth]{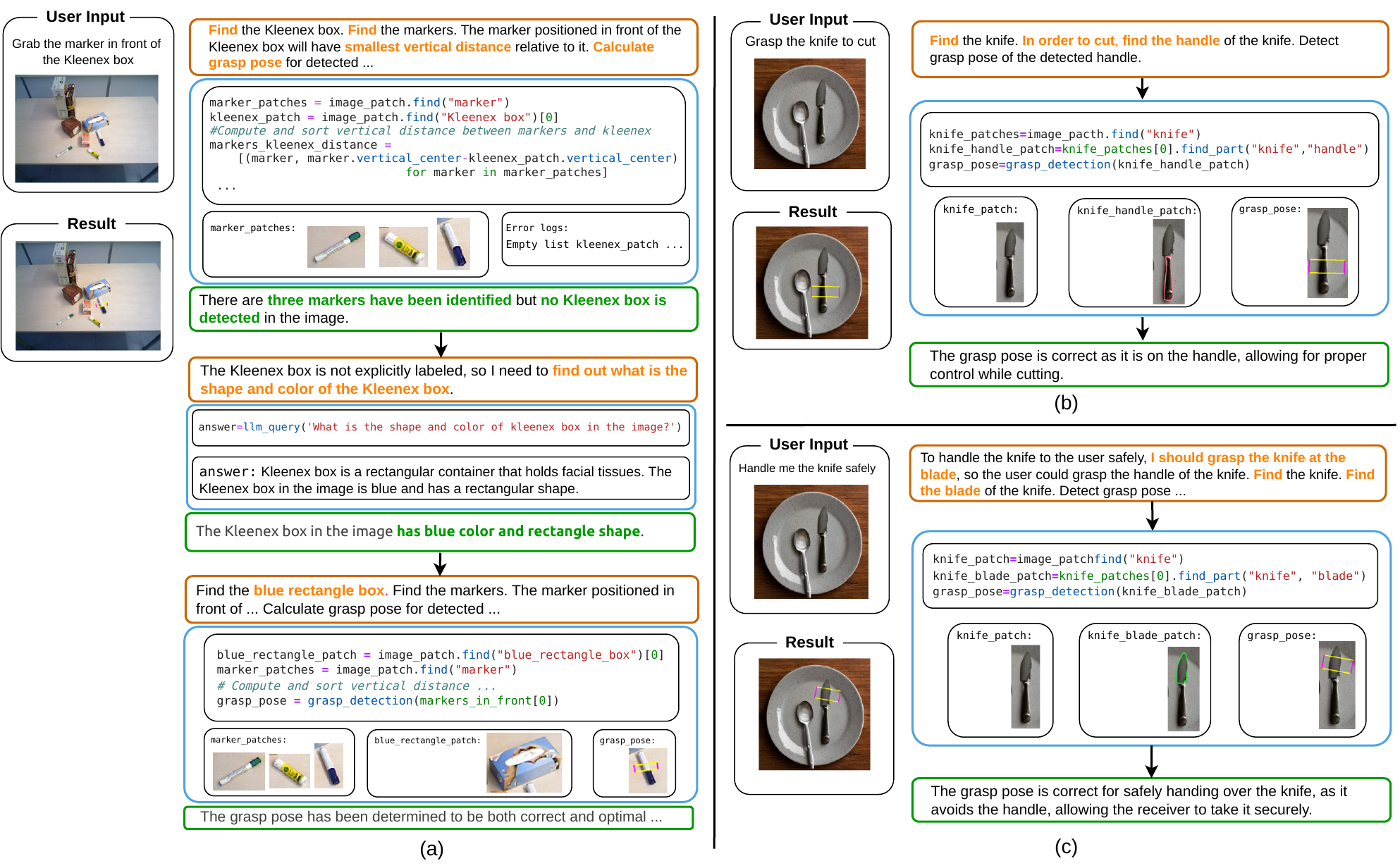}
    \vspace{-1ex}
    \caption{In-the-wild results. This figure illustrates the detailed processing steps of our method, including the planning process, code generation, execution results, and evaluation after each step. Our approach effectively handles ambiguous text queries (a) and adapts to task-oriented grasp detection (b), (c), demonstrating its robustness and flexibility.}
    \label{fig: inthewild}
\end{figure*}

\textbf{In the wild results.} Fig.~\ref{fig: inthewild} shows additional capabilities of our method through real-world examples. Our approach effectively handles ambiguous text queries (Fig.~\ref{fig: inthewild}~(a)). For instance, when a brand name such as ``Kleenex" appears in the query, the model leverages external world knowledge to reinterpret the input. Instead of relying solely on the ambiguous term, it generates a more explicit prompt (``blue box"), enhancing tool usage and reasoning. Additionally, our method demonstrates task-oriented grasp detection, where the generated grasp pose aligns with the action specified in the text prompt (Fig.~\ref{fig: inthewild} (b, c)), ensuring contextually appropriate grasping. These examples demonstrate that our method is highly adaptable and capable of handling a wide range of grasp detection tasks.

\begin{table}[t]
\caption{Comparison with supervised methods}
\vspace{4pt}
\centering
\renewcommand\tabcolsep{8pt} 
\renewcommand\arraystretch{1} 
\begin{tabular}{lcc}
\toprule
\textbf{Baseline} & \textbf{Success rate} & \textbf{Inference (s)} \\ \midrule
CLIP-Fusion~\cite{xu2023joint} & 0.33 \textcolor{OliveGreen}{(+0.35)} & $0.157^{\pm 0.002}$\\
CLIPORT\cite{shridhar2022cliport} & 0.29 \textcolor{OliveGreen}{(+0.39)} & $0.131^{\pm 0.002}$\\
LGD~\cite{vuong2024language} & 0.45 \textcolor{OliveGreen}{(+0.23)} & $22.00^{\pm 0.250}$\\
MaskGrasp~\cite{vo2024language} & 0.45 \textcolor{OliveGreen}{(+0.23)} & $0.116^{\pm 0.003}$\\
LLGD~\cite{nguyen2024lightweight} & 0.46 \textcolor{OliveGreen}{(+0.22)} & $0.264^{\pm 0.020}$\\
GraspMamba~\cite{nguyen2024graspmamba} & 0.52 \textcolor{OliveGreen}{(+0.16)} & $\mathbf{0.032^{\pm 0.001}}$ \\ 
GraspSAM~\cite{noh2024graspsam} & 0.63 \textcolor{OliveGreen}{(+0.05)} & $0.510^{\pm 0.007}$\\\midrule
GraspMAS (ours) & \textbf{0.68} & $2.120^{\pm 0.500}$\\ \bottomrule
\end{tabular}
\label{tab: compare_sup}
\end{table}

\subsection{Robot Experiments in Simulation}

\textbf{Setup.} Prior works~\cite{vuong2024language, vo2024language, nguyen2024graspmamba} have validated language-driven grasp results on real robots. While this is essential, it may not be feasible for other research groups with limited hardware access. To tackle this, we setup a simulation environment using the Maniskill simulator~\cite{taomaniskill3} with a Franka Emika Panda robot operating in a tabletop setting. We conduct experiments on two scenarios: in the single scenario, only one object is present; in the clutter scenario, between 5 and 15 objects are arranged randomly. We use 65 simulated objects from the YCB dataset~\cite{calli2015benchmarking}. Each scenario comprises 50 samples paired with a corresponding text prompt. 

\begin{table}[t]
\caption{Robotic grasp detection results in simulation}
\centering
\vspace{10pt}
\begin{tabular}{lccl}
\toprule
\textbf{Method} & \textbf{Single} & \textbf{Clutter} \\
\midrule
OWLv2~\cite{minderer2024scaling} + RAGT~\cite{cao2023nbmod} & 0.40 \textcolor{OliveGreen}{(+0.42)} & 0.18 \textcolor{OliveGreen}{(+0.54)} \\
GroundDINO~\cite{liu2024grounding} + RAGT~\cite{cao2023nbmod} & 0.42 \textcolor{OliveGreen}{(+0.40)} & 0.22 \textcolor{OliveGreen}{(+0.50)} \\
QWEN2-VL~\cite{wang2024qwen2} + RAGT~\cite{cao2023nbmod} & 0.62 \textcolor{OliveGreen}{(+0.20)} & 0.36 \textcolor{OliveGreen}{(+0.36)} \\
OWG~\cite{tziafas2024towards} & 0.66 \textcolor{OliveGreen}{(+0.16)} & 0.58 \textcolor{OliveGreen}{(+0.14)} \\
Viper-GPT~\cite{suris2023vipergpt} & 0.70 \textcolor{OliveGreen}{(+0.12)} & 0.62 \textcolor{OliveGreen}{(+0.10)} \\
GraspMAS (ours) & \textbf{0.82} & \textbf{0.72} \\ \bottomrule
\end{tabular}
\label{tab: simulation}
\end{table}

\textbf{Results.} Table~\ref{tab: simulation} shows the success rate of our method and other baselines in the simulation experiment. The results demonstrate that our method outperforms baseline approaches. This highlights the strong zero-shot capabilities of our approach to simulation environments. Fig.~\ref{fig: robot} (a) further illustrates a successful case. More illustrations can be found in our Supplementary Video.

\subsection{Real Robot Experiment}
\begin{figure}[t]
    \centering
    \includegraphics[width=0.98\linewidth]{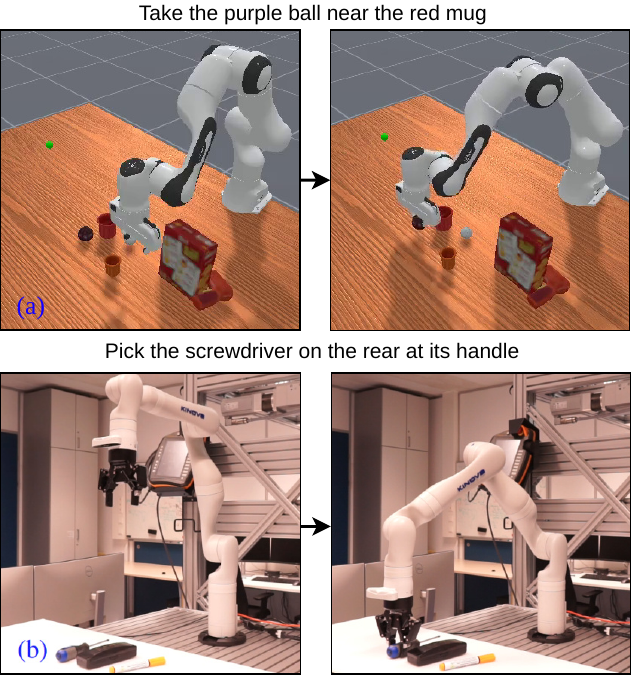}
    \vspace{0.7ex}
    \caption{Robotic demonstration. (a) Simulation environment with Franka Emika robot, (b) Real-world experiment with Kinova Gen 3 robot.
    }
    \label{fig: robot}
\end{figure}

\textbf{Setup.} Fig.~\ref{fig: robot} (b) illustrates the evaluation of our method with a Kinova Gen3 7-DoF robot. We utilize RGB data acquired from an Intel RealSense D410 camera to perform grasp detection, alongside other techniques outlined in Table~\ref{tab:real_robot}. Our approach first estimates 5-DoF grasp poses, which are then transformed into 6-DoF poses under the assumption that objects rest on flat surfaces. The robot is guided toward these poses through trajectory optimization methods~\cite{vu2023machine}. Both inference and control are managed on a computer with an Intel Core i7 12700K processor and an NVIDIA RTX 4080S Ti graphics card. We conducted tests in scenarios involving both single objects and cluttered arrangements, using a range of real-world items, and each experiment was repeated 25 times to ensure consistent comparisons across all methods. The same text prompts are used for all methods in each experiment setup for a fair comparison. 

\textbf{Results.} Table~\ref{tab:real_robot} show the results of the real robot grasping experiment. We compare our method with the state-the-art end-to-end training methods, including LGD~\cite{vuong2024language}, MaskGrasp~\cite{vo2024language}, GraspMamba~\cite{nguyen2024graspmamba}, and compositional methods (OWG~\cite{tziafas2024towards} and Viper-GPT~\cite{suris2023vipergpt}). The results show that our GraspMAS significantly surpasses the performance of other baselines in both the single and cluttered setup. Overall, we observe that the compositional approaches, such as our method and Viper-GPT~\cite{suris2023vipergpt} achieve better results than the end-to-end methods such as GraspMamba~\cite{nguyen2024graspmamba}.

\begin{table}[t]
\caption{Results with Real Robot Experiments}
\centering
\vspace{10pt}
\begin{tabular}{lccl}
\toprule
\textbf{Method} & \textbf{Single} & \textbf{Clutter} \\
\midrule
LGD~\cite{vuong2024language} & 0.43 \textcolor{OliveGreen}{(+0.37)} & 0.42 \textcolor{OliveGreen}{(+0.34)} \\
MaskGrasp~\cite{vo2024language} & 0.42 \textcolor{OliveGreen}{(+0.38)} & 0.42 \textcolor{OliveGreen}{(+0.34)} \\
GraspSAM~\cite{noh2024graspsam} & 0.45 \textcolor{OliveGreen}{(+0.35)} & 0.43 \textcolor{OliveGreen}{(+0.33)} \\
GraspMamba~\cite{nguyen2024graspmamba} & 0.54 \textcolor{OliveGreen}{(+0.26)} & 0.52 \textcolor{OliveGreen}{(+0.24)} \\ 
 \midrule
OWG~\cite{tziafas2024towards} & 0.52 \textcolor{OliveGreen}{(+0.28)} & 0.48 \textcolor{OliveGreen}{(+0.28)}  \\
Viper-GPT~\cite{suris2023vipergpt} & 0.64 \textcolor{OliveGreen}{(+0.16)} & 0.52 \textcolor{OliveGreen}{(+0.24)} \\
GraspMAS (ours) & \textbf{0.80} & \textbf{0.76}\\
\bottomrule
\end{tabular}
\label{tab:real_robot}
\end{table}

\begin{figure}[t]
    \centering
    \includegraphics[width=\linewidth]{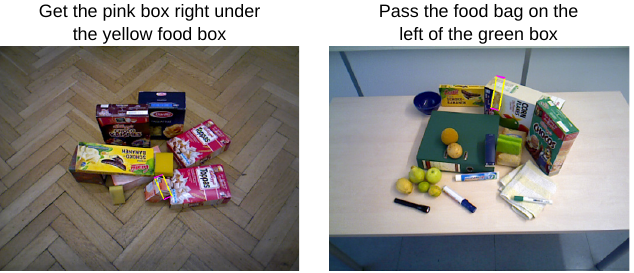}
    \vspace{-1ex}
    \caption{Failure cases of our method.}
    \label{fig: failure_case}
\end{figure}
\subsection{Discussion}  
\textbf{Practical application.} Through intensive experiments, our GraspMAS shows a strong reasoning ability through its compositional design, which allows the robots to interpret complex human commands and deliver precise inferences. Furthermore, as GraspMAS is a compositional approach, it enables seamless integration of new models (as tools) from the literature, so improvements of these tools directly enhance the overall performance of our pipeline. Our method is especially advantageous in complex and dynamic human-robot interaction scenarios and unseen environments.

\textbf{Limitations.} While our method effectively resolves complex text queries and significantly improves accuracy, it still has certain limitations. 
First, its slow inference makes it less suitable for industrial or time-critical applications where rapid responses are essential. To address this, we could deploy lightweight tool alternatives to reduce computational load. Second, our method encounters difficulties in densely cluttered environments where objects significantly overlap, as demonstrated in Fig.~\ref{fig: failure_case}. To overcome this, we could explore a manipulation framework that enables the robot to remove obstacles, thereby resolving ambiguities.

\section{Conclusions}\label{Sec:con}
We introduce GraspMAS, a new multi-agent framework for language-driven grasp detection. Our approach leverages foundation models as tools to interpret complex queries and understand spatial relationships. Through the closed-loop interaction of three agents \texttt{Planner}, \texttt{Coder}, and \texttt{Observer}, our method can dynamically reason in diverse real-world scenarios, ensuring effective and context-aware grasping. The experiment results on GraspAnything++ and OCID-VLG datasets demonstrate the significant improvements of our method compared to baselines. The effectiveness of our method is further validated through experiments in both simulation and real-world robotic settings. Our code will be released to encourage future study.


\bibliographystyle{class/IEEEtran}
\bibliography{class/IEEEabrv,class/reference}
   
\end{document}